\documentclass[conference]{IEEEtran}
\IEEEoverridecommandlockouts
\usepackage{cite}
\usepackage{amsmath,amssymb,amsfonts}
\usepackage{bm}
\usepackage{algorithmic}
\usepackage{graphicx}
\usepackage{textcomp}
\usepackage{booktabs}
\usepackage[small]{caption}
\usepackage{subcaption}
\usepackage[dvipsnames,table,xcdraw]{xcolor}
\usepackage{adjustbox}
\usepackage{graphicx}
\usepackage{listings}
\usepackage{hyperref}
\usepackage{todonotes}
\graphicspath{{../}{./}}

\hypersetup{
    colorlinks=true,
    urlcolor=MidnightBlue,
    linkcolor=MidnightBlue,
    citecolor=MidnightBlue,
}

\definecolor{codegreen}{rgb}{0,0.6,0}
\definecolor{codegray}{rgb}{0.5,0.5,0.5}
\definecolor{codepurple}{rgb}{0.58,0,0.82}
\definecolor{backcolour}{rgb}{0.95,0.95,0.92}
\definecolor{summarygray}{rgb}{0.5, 0.5, 0.5}

\lstdefinestyle{mystyle}{
    backgroundcolor=\color{backcolour},   
    commentstyle=\color{codegreen},
    keywordstyle=\color{magenta},
    numberstyle=\tiny\color{codegray},
    stringstyle=\color{codepurple},
    basicstyle=\ttfamily\footnotesize,
    breakatwhitespace=false,         
    breaklines=true,                 
    captionpos=b,                    
    keepspaces=true,                 
    numbers=left,                    
    numbersep=5pt,                  
    showspaces=false,                
    showstringspaces=false,
    showtabs=false,                  
    tabsize=2
}

\newcommand{\summaryline}[1]{}

\DeclareMathOperator{\enc}{enc}
\DeclareMathOperator{\dec}{dec}

\DeclareMathOperator{\similarity}{sim}
\DeclareMathOperator{\diversity}{diversity}

\def\BibTeX{{\rm B\kern-.05em{\sc i\kern-.025em b}\kern-.08em
    T\kern-.1667em\lower.7ex\hbox{E}\kern-.125emX}}

\begin{document}
\title{Mario Plays on a Manifold: Generating Functional Content in Latent Space through Differential Geometry
}

 \author{\IEEEauthorblockN{Miguel González-Duque$^{1}$, Rasmus Berg Palm, Søren Hauberg$^{2}$, Sebastian Risi$^{1, 3}$}
 \IEEEauthorblockA{\textit{$^{1}$Creative AI Lab, IT University of Copenhagen} \\
 \textit{$^{2}$Cognitive Systems, Technical University of Denmark}\\
  \textit{$^{3}$modl.ai Denmark}\\
 \texttt{migd@itu.dk, rasmusbergpalm@gmail.com, sohau@dtu.dk, sebr@itu.dk}}}




\maketitle

\begin{abstract}
    \summaryline{The problem: decoding to functional content in games.}
    Deep generative models can automatically create content of diverse types. However, there are no guarantees that such content will satisfy the criteria necessary to present it to end-users and be functional, e.g.\ the generated levels could be unsolvable or incoherent. In this paper we study this problem from a geometric perspective, and provide a method for reliable interpolation and random walks in the latent spaces of Categorical VAEs based on Riemannian geometry. We test our method with ``Super Mario Bros'' and ``The Legend of Zelda'' levels, and against simpler baselines inspired by current practice. Results show that the geometry we propose is better able to interpolate and sample, reliably staying closer to parts of the latent space that decode to playable content.
    \summaryline{In this paper, we map functional parts of the latent space using GPCs, we then explore this using A* and geometric diffusion.} \summaryline{Results show that we're able to reliably present functional and diverse content, beating baselines.}
\end{abstract}


\section{Introduction}

\begin{figure}
        \centering
        \includegraphics[width=0.95\columnwidth]{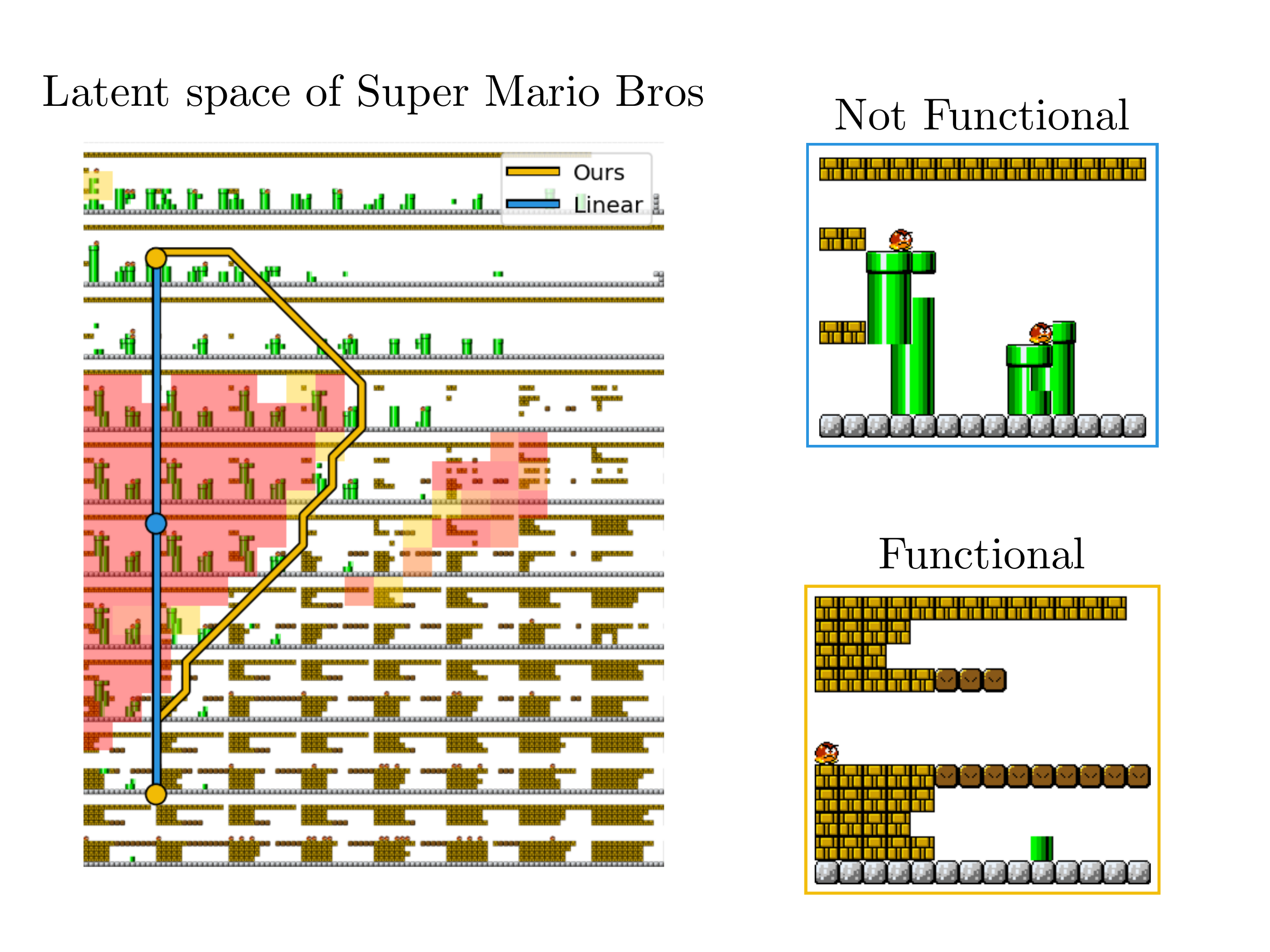}
        
        \caption{\textbf{Functional content in latent space.} When using deep generative models, there are no guarantees that functional content will be produced. This figure shows the result of decoding a grid of levels in the latent space of a Variational Autoencoder, highlighting a functional level (i.e. one solvable by an AI agent) and a non-functional one. Superimposed to the levels is a finer grid showing which regions of the latent space correspond to functional (transparent) and non-functional (red) content. We propose a method for performing interpolations and random walks that stay within functional content. This figure shows a comparison between an interpolation using the proposed approach, and a linear interpolation.}
        \label{fig:banner}
        \vspace{-0.65cm}
    \end{figure}

    \summaryline{The problem in context.}
    Deep latent variable models, such as Variational Autoencoders (VAEs) or Generative Adversarial Networks (GANs), learn a low-dimensional reprensentation of a given dataset.
    Such methods have found great application in the game AI community, mainly to provide a continuous space in which to search for relevant content using evolutionary techniques or random sampling \cite{giacomello2018doom,MarioGAN,CPPN2GAN}, or as a tool for game blending \cite{sarkar2020conditional,sarkar2021generating,sarkar2021dungeon,yang2020game}. Even though these methods excel at replicating a given distribution and generating novel samples, they sometimes fail to generate \textbf{functional} content \cite{PCGML}. For example, when using them to create levels for tile-based games, there are no guarantees for the generated content to be solvable by players \cite{MarioGAN}, restricting the possibility to serve content directly from latent space. Fig.~\ref{fig:banner} shows an example of this problem for the latent space of a VAE trained on Super Mario Bros (SMB) levels:
    only some of the regions of the latent space correspond to functional levels (defined in this case as levels that are solvable for a human player, or an artificial agent), thus making it difficult to reliably sample or interpolate functional content. Fig.~\ref{fig:banner} shows, for example, a linear interpolation that crosses non-functional regions of latent space.

    \summaryline{This problem, framed as a geometrical one: building a metric that penalizes non-playability.}
    We propose a heuristic for safe interpolation and random walks
    based on differential geometry. Interpolations and random walks allow us to gradually modify one type of content to another, making them great tools for designers interested in creating functional content \cite{schrum2020interactive}. In this novel perspective, we can construct a discrete graph of the points in latent space that correspond to playable content, and perform interpolations and random walks therein. Fig.~\ref{fig:banner} shows an example of our interpolation approach.
    These techniques are inspired by advancements in the uncertainty quantification community, where methods like these are applied to stay close to where training points are \cite{arvanitidis:iclr:2018,Tosi:UAI:2014}. Our insight is that these same tools could be applied for staying close to playable content instead of the support of the data.

    \textbf{High-level description of our method.}~
    As discuss above, we provide a method for interpolating and performing random walks based on considering a discrete graph of only the playable content in latent space. To construct it, we first train a Variational Autoencoder with one hierarchical layer on tile-based game content (e.g.\ SMB levels), and we regularize its decoder to be uncertain in non-playable regions. This regularization process, developed for differential geometry, assigns high volume to non-functional parts of latent space, thus forcing shortest paths and random walks to avoid them. As inputs, our method requires a Variational Autoencoder (with one hierarchical layer and low-dimensional latent space) trained on tile-based content, and a way to test for the functionality of decoded content.

    \summaryline{Short summary of contributions and results. Check the methods and results sections to know what to summarize}
    We test our method against simpler baselines (two types of random walks based on taking Gaussian steps and on following the mass of playable levels, and linear interpolation), and we find that we are able to present functional content more reliably than them. These methods are tested on two domains: grid-based representations of levels from Super Mario Bros and The Legend of Zelda, using different definitions of functionality (e.g.\ a solvable level, or a solvable level involving jumps).
    

    \section{Background}

    \summaryline{The method's big picture: having a DGM on some content, identifying the regions of interest, using geometry to explore them.} From a bird eye's view, our method can be described by the following steps: (i) train a Variational Autoencoder (VAE) on tile-based data (using a low-dimensional latent space), (ii) explore its latent space for functional content using grid searches, and (iii) use this knowledge to build a discrete graph that connects only the functional content. To construct this graph, we leverage differential geometry: we modify the decoder such that non-playable areas have high volume. This section describes all the theory and technical tools involved. 

    \subsection{Variational Autoencoders on tile-based data}

    \begin{figure}
        \centering
        \includegraphics[width=0.85\columnwidth]{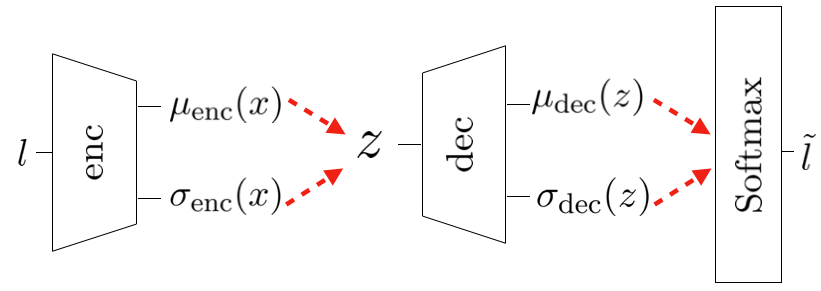}
        \caption{\textbf{Description of the VAE.} In our experiments, we use a one-layer hierarchical VAE. This means that we first decode to a Normal distribution, whose samples are then passed through a Softmax activation function to generate levels. Red dashed arrows indicate sampling from a normal distribution.}
        \label{fig:model:description_of_VAE}
        \vspace{-0.35cm}
    \end{figure}
    
    \summaryline{What are VAEs.} Variational Autoencoders (VAEs) \cite{Kingma2014:VAE,pmlr-v32-rezende14} are a type of latent variable deep generative model (i.e.\ a method for approximating the density $p(x)$ of a given dataset, assuming that the generative process involves a low dimensional representation given by latent codes $z$). Opposite to GANs~\cite{GANs}, VAEs model this density explicitly using a variational training process. This means that the conditional distribution $p(x|z)$ (which models the probability density of a certain datapoint $x$ given a certain training code $z$) and the posterior distribution of the training codes $p(z|x)$ (which models the likelihood of the code of a certain datapoint) are approximated using parametrized families of distributions. A common choice is to approximate the posterior density of the training codes using a Gaussian distribution $q_\phi(z|x)$, parametrized using neural networks. Since we will focus on tile-based content, the conditional $p_\theta(x|z)$ will be modeled using a Categorical distribution parametrized by a neural network.

    In practical terms, this translates to having an autoencoder with \textit{probabilistic} encoder and decoder, both learning the parameters of the distributions $q_\phi(z|x)$ and $p_\theta(x|z)$ respectively.
    
    \summaryline{Describe our specific model.} For our geometric tools to work, we need a deep generative model that allows for manipulating the uncertainty of our decoded outcomes. Thus, we consider a one-layer hierarchical VAE which first decodes to a Normal distribuition, which is then sampled to create logits for the final Categorical distribution $p(x|z)$. This hierarchical layer computes the mean $\mu_{\dec}(z)$ and standard deviations $\sigma_{\dec}(z)$ of the logits of the Categorical distribution. In other words, the probabilities of this distribution are giving by
    \begin{equation}\label{eq:model:decoder_w_softmax}
        \text{decode}(z) = \text{Softmax}(\mu_{\text{dec}}(z) + \varepsilon\odot\sigma_{\text{dec}}(z)^2),
    \end{equation} where $\epsilon\sim N(0, I_D)$ and $D$ is the dimension of the data. With this formulation, we can manipulate the uncertainty of our Categorical by having $\sigma_{\dec}(z) \to \infty$. A diagram that shows how this network is structured can be found in Fig.~\ref{fig:model:description_of_VAE}.

    Summarizing, our model is as follows
    \begin{align}
        q(z\,|\,x) &= \mathcal{N}\left(z \,|\, \mu_\text{enc}(x), \sigma_\text{enc}(x)^2\right), \\
        p(x\,|\,z) &= \text{Cat}(x \,|\, \text{decode}(z)),
    \end{align}
    where $\mu_{\enc}$, $\mu_{\dec}$, $\sigma_{\enc}$ and $\sigma_{\dec}$ are usually parametrized using neural networks whose weights are optimized by maximizing the Evidence Lower Bound (ELBO) \cite{Kingma2014:VAE}.

    \begin{figure}
        \centering
        \begin{subfigure}[b]{0.5\columnwidth}
            \centering
            \includegraphics[width=0.95\textwidth]{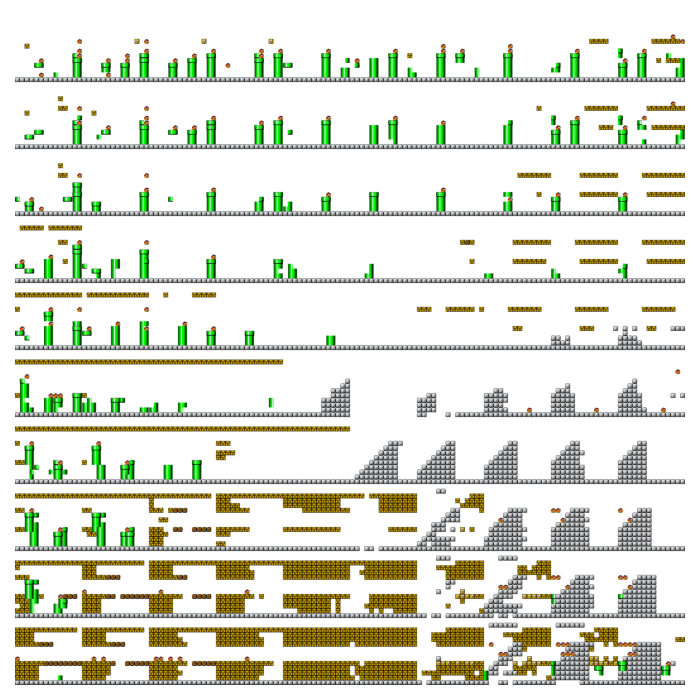}
            \caption{}
            \label{fig:model:grid_in_latent_space}
        \end{subfigure}%
        \begin{subfigure}[b]{0.5\columnwidth}
            \centering
            \includegraphics[width=0.95\textwidth]{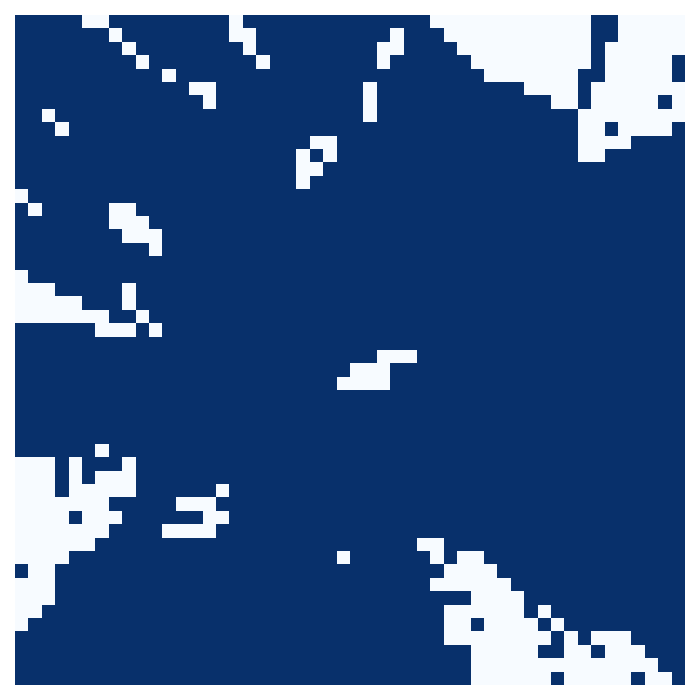}
            \caption{}
            \label{fig:model:playability_in_latent_space}
        \end{subfigure}
        
        \caption{\textbf{How playability is distributed in the latent space of SMB.} After training the VAE, we decode in levels in a $10\times10$ grid of equally spaced points in the $[-5,5]\times[-5,5]$ square (Fig.~\ref{fig:model:grid_in_latent_space}). Even in two dimensions, our model is able to segment functionally different levels in distant parts of the latent space. These levels seem to vary smoothly with respect to playability: Fig.~\ref{fig:model:playability_in_latent_space} shows the measured playability after decoding a $50\times 50$ grid of levels in latent space. Blue squares correspond to playable levels, and white to non-playable.}
        \label{fig:model:latent_space}
        \vspace{-0.35cm}
    \end{figure}
    
    \subsection{An intuitive introduction to differential geometry}
    \label{sec:diff_geo}

    \summaryline{A primer on Riemannian manifolds from the intuition} 
    
    
    Our method relies on the theory of smooth and Riemannian manifolds (known as differential geometry). We will explain it by giving an intuition for what Riemannian manifolds are, what a metric is, and how a smooth map can allow us to ``pull back'' a metric from one Riemannian manifold to another. This explanation focuses on intuition at the expense of technical precision. For a thorough and theoretical explanation we recommend \cite{Lee00:introduction_smooth_manifolds,Lee18:introduction_riemannian_manifolds}.
    
    Intuitively, a Riemannian manifold is a smooth surface that carries a \textbf{metric}: a tool for defining distances accounting for how curved the surface might be. For example, distances along a sphere are different from distances on Euclidean space. In loose terms, a metric is equivalent to a dot product, allowing us to define norms (e.g.\ $\|x\| = \sqrt{x\cdot x}$ in $\mathbb{R}^n$), lengths and angles between tangent vectors to the surface. One key subtlety is that this metric is defined locally around a given point, instead of globally as in $\mathbb{R}^n$.

    Consider our decoder $\dec$ between the latent space $\mathcal{Z}$ and the Euclidean data space $\mathbb{R}^D$. This decoder induces a metric on $\mathcal{Z}$ called the \textbf{pullback}, which is given by $M(z) = J_{\dec}(z)^\top J_{\dec}(z)$ where $J_{\dec}$ is the Jacobian of the decoder. To understand this definition in context, we can compute a curve's length in $\mathcal{Z}$ by using lengths in $\mathbb{R}^D$: let $c:[0, 1]\to\mathcal{Z}$ be a curve, its length is then given by
    \begin{align*}
        \text{Length}[c] &= \int_0^1 \left\|\frac{\mathrm{d}}{\mathrm{d}t}d(c(t))\right\| \mathrm{d}t \\
        &= \int_0^1 \sqrt{(J_d(c(t)) c'(t))\cdot (J_d(c(t)) c'(t))} \mathrm{d}t \\
        &= \int_0^1 \sqrt{c'(t)^\top M(c(t)) c'(t)} \mathrm{d}t
    \end{align*}
    where we used the chain rule and the definition of the usual norm in Euclidean space.
    In other words, the pullback $M(z)$ is mediating local lengths in $\mathcal{Z}$. The quantity $\det(M(z))$ also plays a special role, since it measures the \textbf{volume} of a region locally around $z$ \cite{Lee18:introduction_riemannian_manifolds}. Our method modifies the decoder in such a way that $M(z)$ has high volume close to non-playable parts of latent space.

    \section{Defining a geometry in latent space}
    \label{sec:discretized_geometry}
    
    \subsection{Playability-induced geometry in latent space: Motivation}
    \label{sec:motivation_for_geometry}


    \begin{figure*}
        \centering
        \begin{subfigure}[b]{0.185\textwidth}
            \centering
            \includegraphics[width=1.0\textwidth]{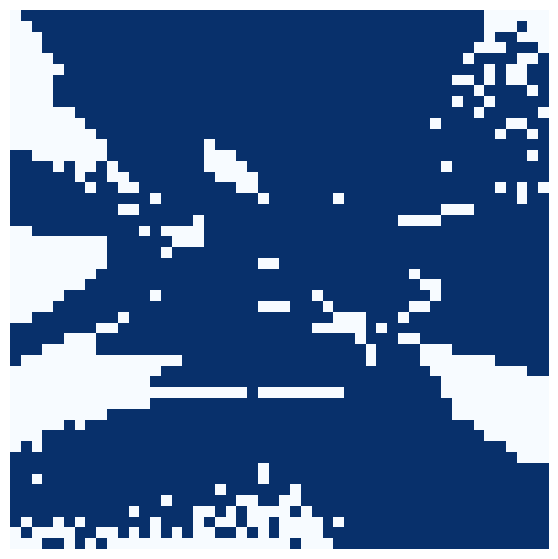}
            \caption{Coherent levels}
            \label{fig:approximation:obstacles}
        \end{subfigure}%
        \begin{subfigure}[b]{0.2\textwidth}
            \centering
            \includegraphics[width=1.0\textwidth]{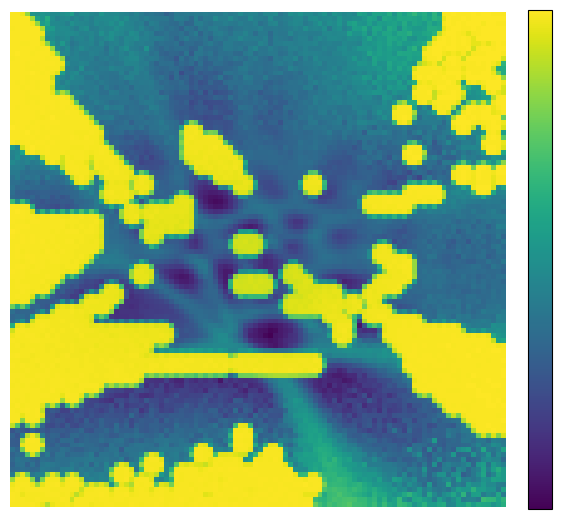}
            \caption{Metric volume}
            \label{fig:approximation:metric_volume}
        \end{subfigure}%
        \begin{subfigure}[b]{0.185\textwidth}
            \centering
            \includegraphics[width=1.0\textwidth]{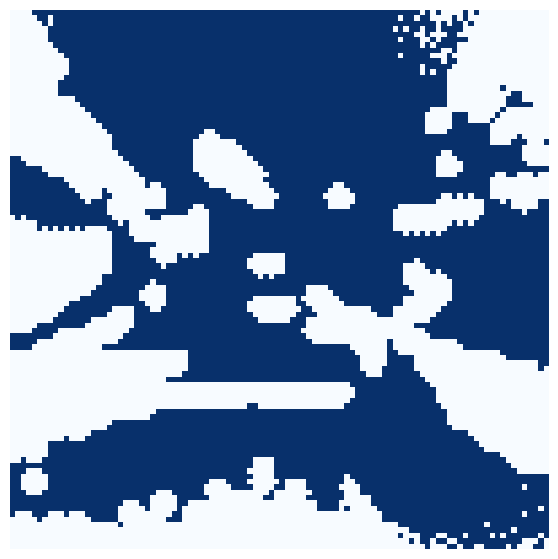}
            \caption{Discrete approximation}
            \label{fig:approximation:approximation}
        \end{subfigure}%
        \begin{subfigure}[b]{0.335\textwidth}
            \centering
            \includegraphics[width=1.0\textwidth]{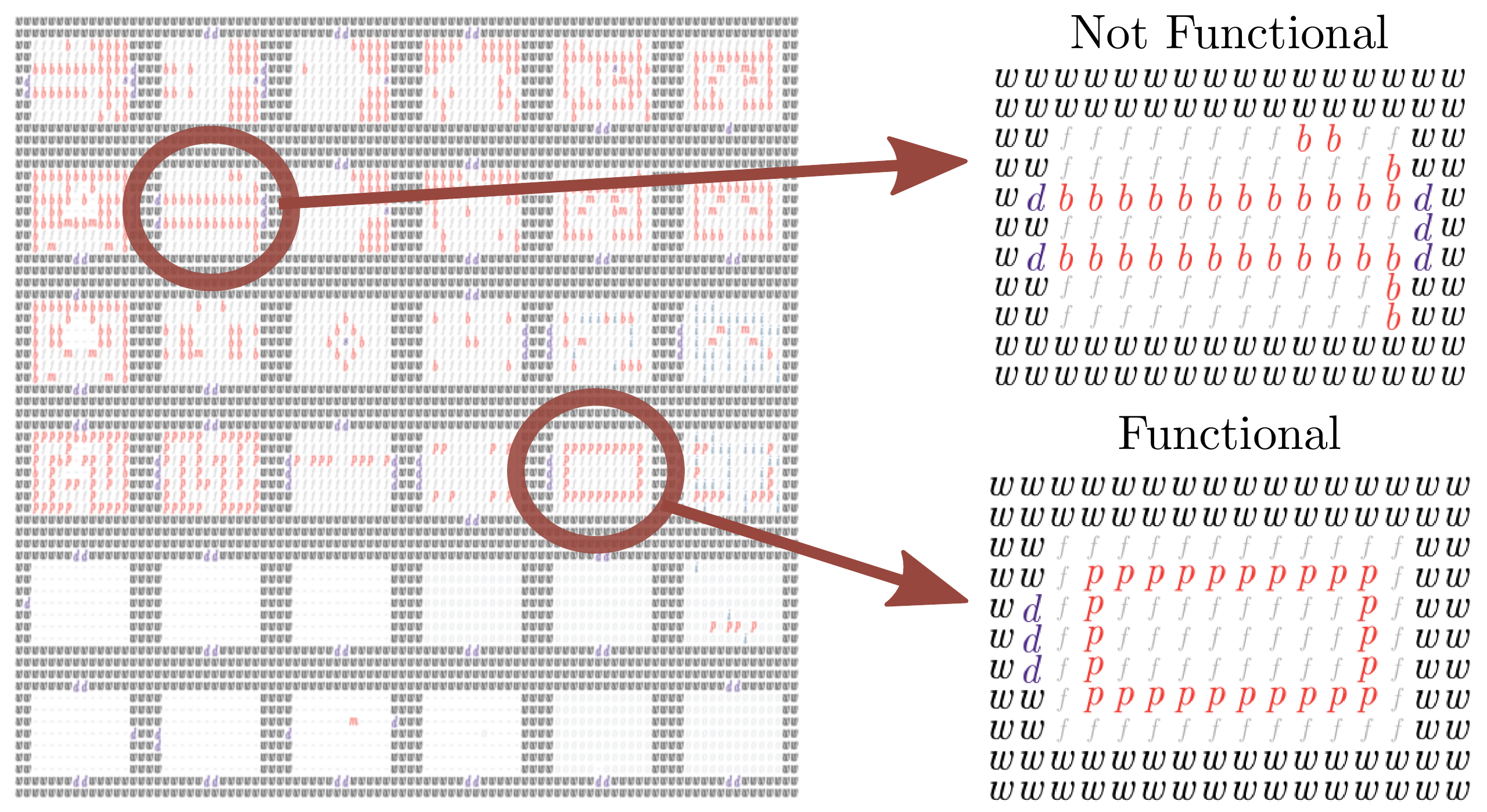}
            \caption{Levels in latent space}
            \label{fig:approximation:grid}
        \end{subfigure}%
        
        \caption{\textbf{Defining a geometry in latent space.} Here we illustrate how our method works using a VAE trained on Zelda levels. First, in Fig.~\ref{fig:approximation:obstacles} we decode a coarse grid in latent space and identify the regions where non-functional content is (blue corresponds to functional). Using this knowledge, we then calibrate a decoder to have high metric volume in these non-functional regions (see Fig.~\ref{fig:approximation:metric_volume}; we show log-volumes). From these metric values, we construct an approximation by thresholding, arriving at the discrete graph presented in Fig.~\ref{fig:approximation:approximation}. Notice how there is a wider band around non-functional content, helping us to avoid it. Finally, we verify by decoding a grid of levels in latent space in Fig.~\ref{fig:approximation:grid}.}
        \label{fig:approximation}
        \vspace{-0.4cm}
    \end{figure*}

    \summaryline{We find, empirically, that there are certain playability regions in latent space.} After training a VAE on SMB levels, we wondered how playability was distributed in the latent space. For this, we used Robin Baumgarten's super-human A* agent \cite{MarioAICompetition:Baumgarten:2010} to simulate a $50\times50$ grid in $[-5, 5]^2$ in latent space. This agent returns telemetrics, including whether the level was playable and how many jumps the agent performed.
    Simulations are supposed to be deterministic, but we found in practice that they were not and thus we performed 5 simulations per level.\footnote{We hypothesize that this is the effect of random CPU allocations during the simulation. After giving it our best efforts, we were not able to force Baumgarten's agent to be deterministic.}

    Fig.~\ref{fig:model:playability_in_latent_space} shows the VAE's latent space illuminated by the mean playability of each level. Notice that unplayable regions are localized in different parts of the latent space, and can be thought of as ``obstacles'' when doing interpolations or random searches. This empirical observation is supported by looking at different training runs. A similar phenomena can be observed in the latent space of Zelda: if we define functionalty as decoding coherent and connected levels, we see structure in the latent space (see Figs~\ref{fig:approximation:obstacles} and \ref{fig:approximation:grid}).

    \summaryline{We then build a fine grid of playability, and use it to define a geometry in which we can do interpolation and diffusion}
    Our goal, then, is to use this playability in latent space (or an approximation thereof) to define a meaningful geometry in latent space that allows us to do informed interpolations and diffusions (i.e. random walks).

    \summaryline{Main idea: define a graph in latent space and use simple algorithms in graphs} This section describes how to use the playability information from a coarse grid in latent space to define algorithms for interpolation and random walks in latent space. On a high-level, this geometry is a discrete approximation of the one induced by $J_{\dec}(z)^\top J_{\dec}(z)$, where $J_{\dec}$ is the Jacobian of a ``calibrated'' version of the decoder.  The quantity $J_{\dec}(z)^\top J_{\dec}(z)$ has a clear geometric meaning: it defines a metric on the latent space $\mathcal{Z}$ by ``pulling back'' the Euclidean metric on the data space $\mathbb{R}^D$ (see Sec.~\ref{sec:diff_geo}). We will guide the reader using a VAE trained on The Legend of Zelda levels as an example (see Fig.~\ref{fig:approximation}).
     



    \subsection{Calibrating the decoder}
    \label{sec:calibrating_the_decoder}
    
    In Sec.~\ref{sec:diff_geo} we saw that we can use the Jacobian of the decoder $\dec$ to define a geometry in the latent space. Let's develop more intuition about what the Jacobian is, and how we can exploit its definition to make traversing through non-playable levels ``more expensive'' for our distances. The Jacobian plays the role of a first order derivative for functions of more than one variable, and it can be approximated by finite differences just like the usual derivative:
    \begin{equation}
        \label{eq:jacobian}
        J_{\dec}(z) \approx \left[\frac{\dec(z + dz_i) - \dec(z)}{\|dz_i\|}\right]_{i=1}^{\dim(\mathcal{Z})}
    \end{equation}
    where $dz$ is a small vector on the $i$th direction and $\|dz_i\|$ is its norm. To increase the volume of the metric (and thus have higher local distances in latent space), we need a decoder in which small changes with respect to the latent code will result in large changes after decoding. Thus, we calibrate our decoder to output what it learned during training when close to playable levels (i.e. $\text{Softmax}(\mu_{\dec}(z) + \epsilon \sigma_{\dec}(z)$), and to extrapolate to noise when close to non-playable levels (replacing $\sigma_{\dec}(z)$ for $10^5$). In other words, whenever we walk close to non-playable levels, we pass through levels of very different forms making the numerator of Eq.~(\ref{eq:jacobian}) explode. This method was first introduced for VAEs in \cite{arvanitidis:iclr:2018} and was applied to Categorical distributions in \cite{detlefsens:proteins:2020}.
    
    Numerically speaking, the way in which we transition from decoding the levels we learned to decoding highly noisy levels is by computing the distance to the closest non-playable level, and transforming that distance into a number $\alpha(z)$ between 0 and 1, where $\alpha(z)=0$ means that the latent code is close to non-playable regions and $\alpha(z)=1$ means the opposite. That number is then used as a semaphore in a sum
    \begin{equation}
        \label{eq:calibrating_the_decoder}
        \dec(z) = \alpha(z)\dec_{\mu, \sigma}(z) + (1 - \alpha(z))\dec_{\mu, 10^5}(z).
    \end{equation}

    This semaphore function $\alpha(z)$ is implemented using a translated sigmoid function. Namely, consider $\text{minDist}(z)$ to be the minimum distance to a non-functional training code, then we can define
    \begin{equation}
        \label{eq:translated_sigmoid}
        \alpha(z) = \text{Sigmoid}\left(\frac{\text{minDist}(z) - \beta k}{\beta}\right),
    \end{equation}
    where $\beta$ is a hyperparameter that governs how quickly we transform from 0 to 1 (we settled for $\beta=5.5$ in our experiments), and $k\approx 6.9$.
    

    In summary, we decode a coarse grid in latent space to identify where possible non-playable regions are, we use it to calibrate our decoder, making it locally noisy around non-playable regions. We then compute the pullback metric of the calibrated decoder, resulting on a notion of distance in which non-playable levels are far away. Figures~\ref{fig:approximation:obstacles} and \ref{fig:approximation:metric_volume} show exactly this: non-functional regions (showed as white) render high log-volume.
    
    \subsection{Approximating the manifold with a graph}
    
    Once we have a calibrated decoder $\dec(z)$, we can equip our latent space with the metric $M(z) = J_{\dec}(z)^\top J_{\dec}(z)$. This metric, as discussed above, has high volume close to non-playable levels.
    Empirically, we found it useful to approximate this manifold using a finite graph. At this point we can easily compute $M(z)$ for many training codes by approximating the Jacobian using first order finite differences (see Eq.~(\ref{eq:jacobian})), knowing that $\det(M(z))$ is a high number for points that are close to non-playable regions of space (according to a coarse grid). We can then sample an arbitrary grid, compute the metric $M(z)$ for each point and construct a finite graph by considering only the points whose volume $\log(\det(M(z))$ is below some threshold $v$ (which we chose to be the mean volume, but can be thought of as a hyperparameter).
    
    Fig.~\ref{fig:approximation:approximation} shows how we can transform a coarse grid into a continuous manifold, and then approximate it using a finite graph as discussed above. Notice that this finite approximation can be of any resolution. In this graph, interpolation and random walks can be easily implemented. We discuss these algorithms next.
    
    \paragraph{Shortest paths on graphs through A*} Once we have a finite graph embedded on $\mathbb{R}^{\dim(\mathcal{Z})}$, computing the shortest path between two latent codes $z$ and $z'$ in the graph can be done using the A* algorithm \cite{OriginalAStar:1968}.
    It must be noted that, for simplicity in the implementation, we consider a connected graph in which the points we are trying to avoid are at infinite distance.
    
    
    \paragraph{Random walks on graphs} Starting at a node $z_0$, our random walk algorithm samples uniformly at random from the set of neighbours of $z_0$, arriving at an intermediate point $z_0^1$. Since the grid is usually very fine (thus making local distances small), we continue this process of sampling uniformly at random from neighbors on $z_0^1$ arriving at $z_0^2$. This process continues for $m$ ``inner'' steps. After these, we assign $z_1 = z_0^m$. For our experiments we decided on $m=25$, but it can be considered a hyperparameter. 
    
    
    \section{Experiments \& Results}
    
    
    \subsection{Training results for the VAE}
    
    \summaryline{Describe the data} We trained our one-layer hierarchical VAEs on data from two tile-based games: Super Mario Bros (SMB) levels and The Legend of Zelda. Both of these are common benchmarks for testing generation in the game AI community. These levels, usually extracted from the Video Game Level Corpus (VGLC)\footnote{\url{https://github.com/TheVGLC/TheVGLC}}, are processed into token-based text files of height and width 14 in the case of SMB, and levels of shape $11\times 16$ in the case of Zelda. This processing involves sliding a $14\times 14$ window across the original SMB levels, and by splitting the original Zelda levels into rooms. We arrive at a total of 2264 playable levels for SMB; and of the 237 original levels for Zelda, we keep 213 that are solvable according to a definition we discuss in the next section.
    
    \summaryline{We trained $n$ different VAEs to prove our point. Describe the hyperparameters and losses.} We trained 10 different VAEs for SMBs and 10 for Zelda. All these networks have MLP encoders with 3 hidden layers of sizes 512, 256 and 128 respectively, then mapping to a latent space of size $2$; the decoders were symmetric, and the last hierarchical layer is of size $D$, the data's size. For the SMB and Zelda networks, we used learning rates of $10^{-3}$ and $10^{-4}$ respectively; for both we used a batch size of 64, and an Adam optimizer \cite{kingma2017adam} and early stopping with at most 250 epochs. All these neural networks were trained using PyTorch \cite{NEURIPS2019_9015}.
    
    We noticed an ELBO loss of $53.4\pm 3.1$ for the training runs on SMB, and of $31.8\pm 10.9$ for Zelda. As can be seen from the variance, the training processes for Zelda were surprisingly unstable, with some networks converging to flat representations, failing to capture the structure of the dataset. Thus, for our experiments we used 4 out of the 10 VAEs trained on Zelda, keeping only the networks in which the latent space had more than flat, constant representations. 

    \subsection{Finding playable regions using a coarse grid}

    \summaryline{On each one of these VAEs we computed the grid of playability. (Mention width of the grid and the setup with the simulator/definition of the grammar for Zelda)} To find the non-playable regions in the latent space of Super Mario Bros, we ran Robin Baumgarten's A* agent \cite{MarioAICompetition:Baumgarten:2010} on the decoded levels of a $50\times 50$ grid in all latent spaces using the simulator provided by the MarioGAN repository\footnote{\url{https://github.com/CIGbalance/DagstuhlGAN}}. In the case of Zelda, we implemented a ``grammar check'' which makes sure that (i) the level has either stairs or doors, (ii) if there are more than two doors, they must be connected by a walkable path (with e.g. no lava/water tiles blocking all possible paths), (iii) doors must be complete and in the right places, and (iv) the level must be surrounded by walls. Figs.~\ref{fig:model:latent_space} and \ref{fig:approximation} show examples of these two latent spaces.

    \begin{figure*}
        \centering
        \begin{subfigure}[b]{0.2\textwidth}
            \centering
            \includegraphics[width=1.0\textwidth]{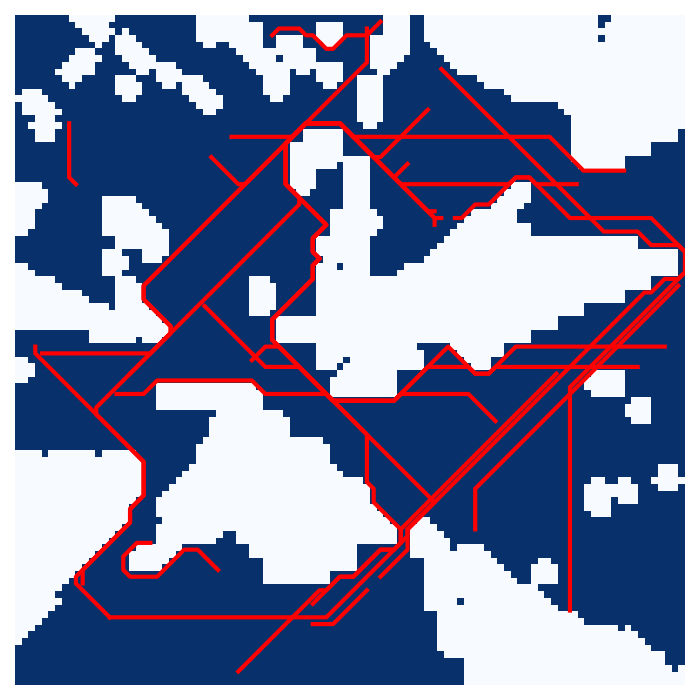}
            \caption{Our interpolations}
            \label{fig:jumping:our_interp}
        \end{subfigure}%
        \begin{subfigure}[b]{0.2\textwidth}
            \centering
            \includegraphics[width=1.0\textwidth]{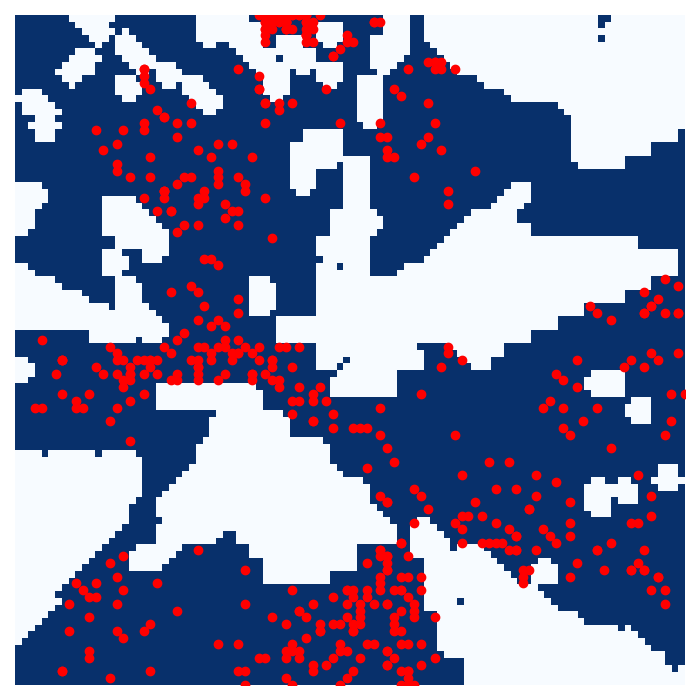}
            \caption{Our random walks}
            \label{fig:jumping:our_diff}
        \end{subfigure}%
        \begin{subfigure}[b]{0.2\textwidth}
            \centering
            \includegraphics[width=1.0\textwidth]{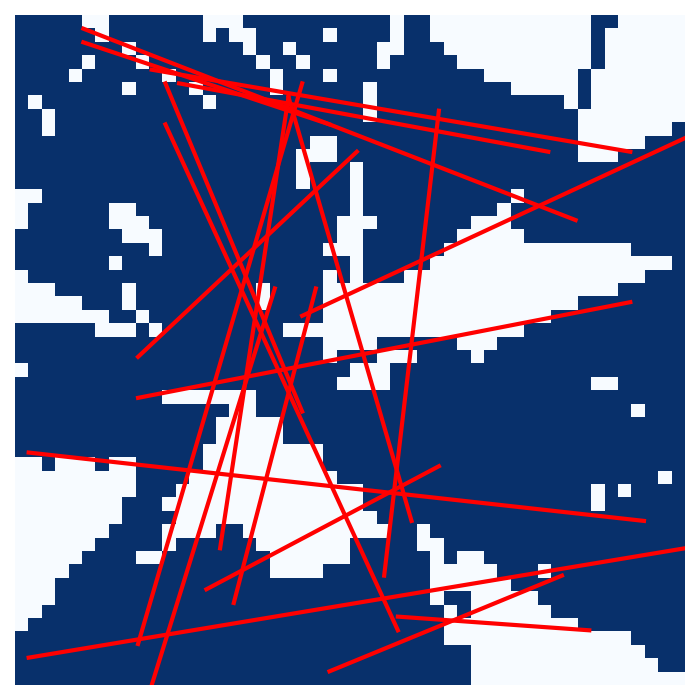}
            \caption{Linear interpolations}
            \label{fig:jumping:linear_interp}
        \end{subfigure}%
        \begin{subfigure}[b]{0.2\textwidth}
            \centering
            \includegraphics[width=1.0\textwidth]{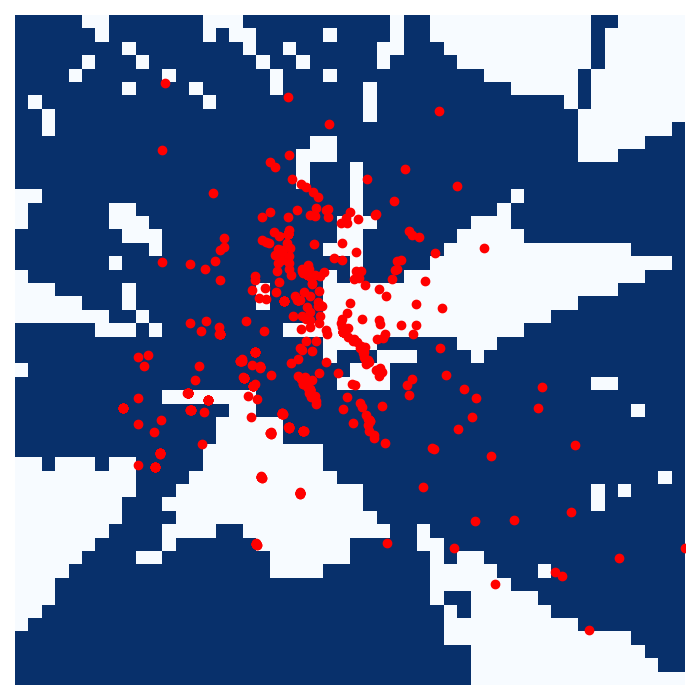}
            \caption{Baseline random walks}
            \label{fig:jumping:baseline_diff}
        \end{subfigure}%
        \begin{subfigure}[b]{0.2\textwidth}
            \centering
            \includegraphics[width=1.0\textwidth]{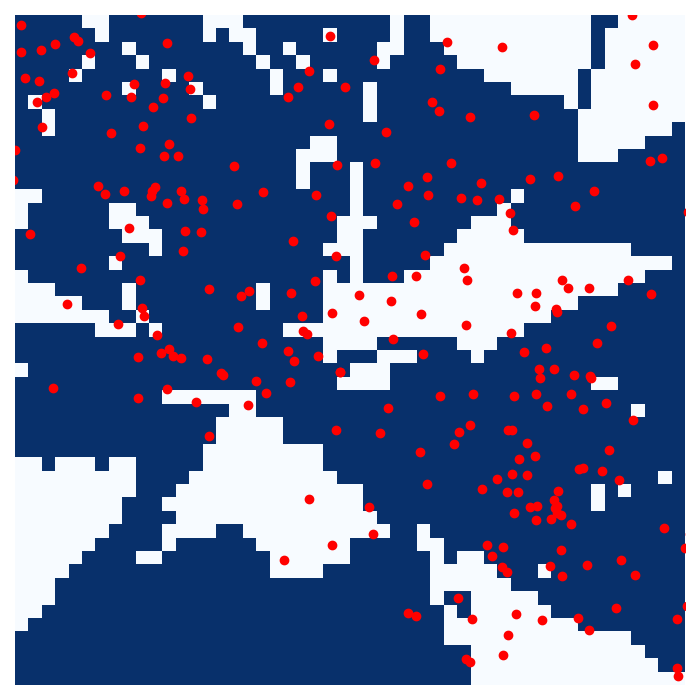}
            \caption{Normal random walks}
            \label{fig:jumping:normal_diff}
        \end{subfigure}%
        
        \caption{\textbf{Interpolations and diffusions in the jumping submanifold.} We showcase examples of the different interpolations and diffusions used in all comparison experiments (see Secs.~\ref{sec:comparing_interpolation_random_walks} and \ref{sec:submanifolds}) for the submanifold given by solvable levels in which Mario jumps. Notice how this graph is a subset of Fig.~\ref{fig:model:playability_in_latent_space}. Figs.~\ref{fig:jumping:our_interp} and \ref{fig:jumping:our_diff} show our interpolations and diffusions respectively. Fig.~\ref{fig:jumping:linear_interp} shows example linear interpolations (used in both baselines) and Figs.~\ref{fig:jumping:baseline_diff} and \ref{fig:jumping:normal_diff} show the random walks of our baselines. We manage to staw away from non-functional levels (shown in white) in both our interpolation and random walks, sometimes at the cost of getting stuck bottlenecks (see the upper part of~\ref{fig:jumping:our_diff}).}
        \label{fig:jumping}
        \vspace{-0.4cm}
    \end{figure*}

    \subsection{Approximating the manifold using graphs}
    
    \summaryline{We can construct graphs of playability of arbitrary sizes. We did it for 100.} After training these VAEs, we can regularize the decoder, forcing it to be expensive in non-playable regions (see Sec.~\ref{sec:calibrating_the_decoder}). When the decoder is calibrated, the log-volume $\log(\det(J_{\dec}(z)^T J_{\dec}(z))) = \log(\det(M(z)))$ is high for latent codes that correspond to non-playable levels.
    
    We can, then, consider a finer grid of any resolution (we chose $100\times 100$ levels), cutting off the nodes for which $\log(\det(M(z))$ is higher than the average over all the grid. This process was used to approximate the playability manifold for both SMB and Zelda. For example, Fig.~\ref{fig:approximation:approximation} shows the resulting $100\times 100$ grid in the Zelda.

    To argue why a discrete approximation was necessary, consider the following random walk algorithm for the continuous case: at any step $z_n$ in the random walk, compute the pullback metric $M(z_n)$ and sample from the Gaussian $N(z_n, M(z_n)^{-1})$. We found, empirically, that these random walks tended to ``get stuck'' close to non-training codes. Indeed, in those regions the inverse of the metric can be thought of as infinitesimally small.
    
    \subsection{Comparing interpolations and random walks}
    \label{sec:comparing_interpolation_random_walks}

    \begin{figure*}
        \centering
        \includegraphics[width=1.9\columnwidth]{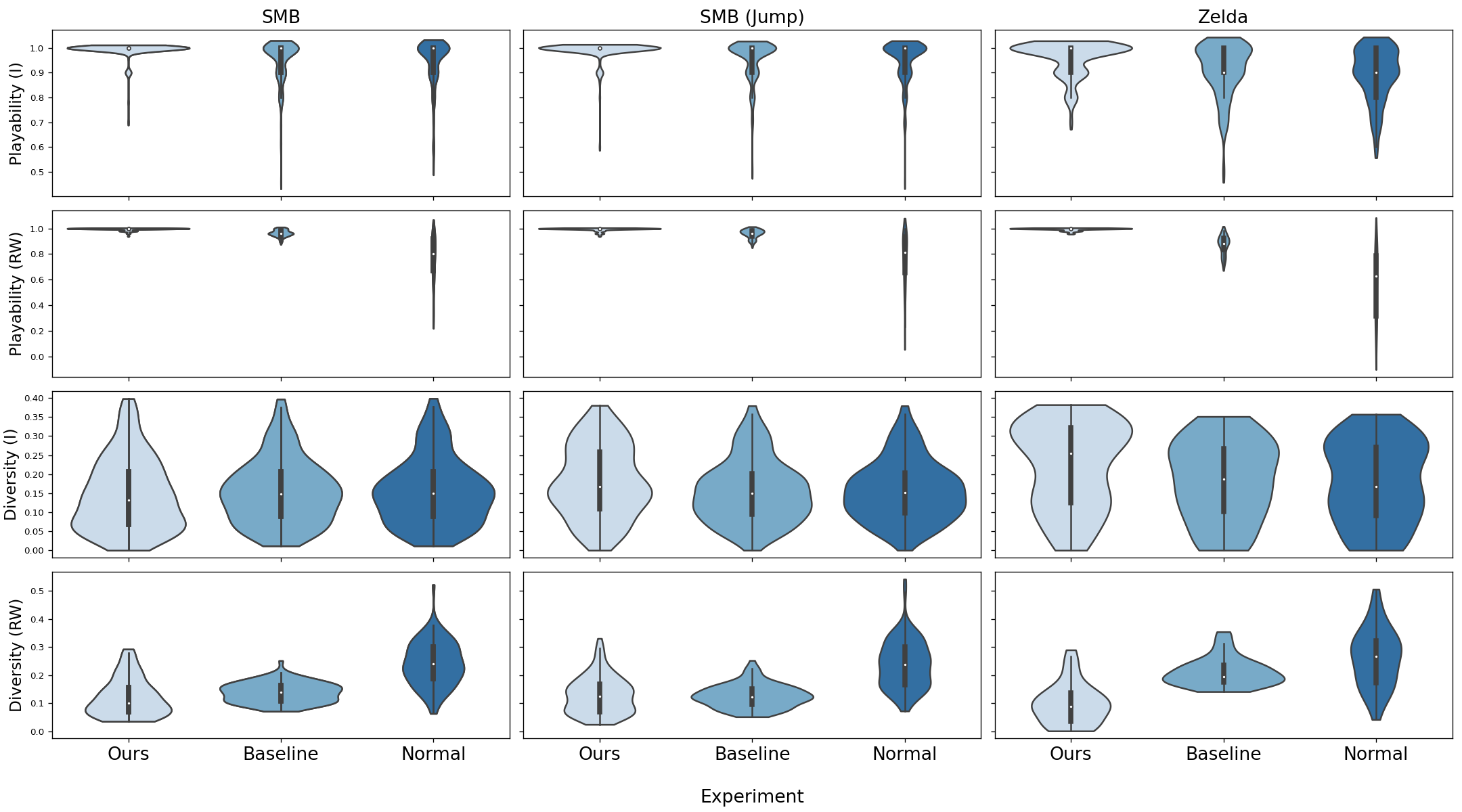}    
        \caption{\textbf{Comparing our geometry against baselines.} This figure shows the distributions of playability and diversity, which are summarized on Table~\ref{table:results:comparing_geometries}, where (I) stands for interpolations and (RW) stands for random walks. For each VAE we performed 20 interpolations and 10 random walks, selecting the starting points at random. These quantities were measured in each interpolation/random walk. Our geometry has most of its playability mass closer to 1.0 than the baselines; however, this comes at a slight cost on diversity: the mass for estimated diversities is lower than the baselines.}
        \label{fig:violin_plots}
        \vspace{-0.4cm}
    \end{figure*} 
    
    \summaryline{Discuss the experimental setup: 10 VAEs, 20 interpolations between random points, 10 diffusions...} To test whether our proposed geometry improved on reliably interpolating/randomly walking between functional content from the latent space we considered, for each VAE, 20 interpolations and 10 random walks. 10 equally spaced points were then selected from each interpolation, and each random walk was ran for 50 steps.

    \summaryline{Baselines} We considered linear interpolation to be a suitable baseline for interpolation, since it is common practice in the deep generative models community \cite{Karras:stylegan:2019}. For random walks, we considered two baselines: first a naïve one, based on randomly sampling at each step $z_n$ from a Gaussian $N(z_n, I_2)$ where $I_2$ is the identity matrix of size 2; secondly a baseline that randomly samples playable levels from the graph and takes a step of fixed size in that direction. This second baseline then tends to stay within playable levels, converging to its center of mass in the limit, and this first baseline is closer to common methods for latent space exploration \cite{MarioGAN}. Example interpolations and random walks for all methods (ours, plus the two baselines) can be found in Figs.~\ref{fig:jumping:linear_interp}, \ref{fig:jumping:normal_diff}, and \ref{fig:jumping:baseline_diff}.

    \summaryline{Discussion of main results for interpolations and random walks} Table~\ref{table:results:comparing_geometries} shows the average playabilities of these 10 interpolations and 20 random diffusions for our proposed geometry, plus the two baselines discussed above. In the case of Super Mario Bros (SMB), we see a slight increase in the reliability of interpolations and random walks: our methods indeed stay more within the playable regions, showing an expected playability (i.e. functionality) of close to $100\%$. This is to be contrasted with e.g.\ the playability of performing Normal random walks, which results only in about $77\%$ of playable levels. In the case of Zelda, we see a similar increase in the probability of presenting a playable level when using our geodesic interpolation; the baselines for random walks, however, seem to perform relatively poorly in these latent spaces. Our method is able to avoid non-playable regions, presenting an estimated $99.5\%$ probability of sampling a random level. Fig.~\ref{fig:violin_plots} shows a violin plot of the mean playabilities for these interpolations and random walks, providing more than a point uncertainty estimate of these results.
    
    \begin{table}
        \centering
        \resizebox{0.99\columnwidth}{!}{
        \begin{tabular}{lllll}
\toprule
{} & \multicolumn{2}{c}{$\mathbb{E}[\text{playability}] \uparrow$} & \multicolumn{2}{c}{$\mathbb{E}[\text{diversity}] \uparrow$} \\
{Geometry} &                    Interpolation &     Random Walks &                  Interpolation &     Random Walks \\
\midrule
                    \multicolumn{5}{c}{Super Mario Bros} \\
                    \midrule
Ours &                  \textbf{0.993}$\pm$0.033 &  \textbf{0.996}$\pm$0.010 &                0.146$\pm$0.034 &  0.121$\pm$0.024 \\
Baseline             &                  0.953$\pm$0.084 &  0.963$\pm$0.026 &                0.154$\pm$0.028 &  0.138$\pm$0.026 \\
Normal               &                  0.949$\pm$0.093 &  0.773$\pm$0.169 &                \textbf{0.155}$\pm$0.029 &  \textbf{0.240}$\pm$0.026 \\
                    \midrule
                    \multicolumn{5}{c}{The Legend of Zelda} \\
                    \midrule
Ours &                  \textbf{0.961}$\pm$0.068 &  \textbf{0.995}$\pm$0.011 &                \textbf{0.222}$\pm$0.112 &  0.099$\pm$0.072 \\
Baseline             &                  0.916$\pm$0.104 &  0.874$\pm$0.073 &                0.182$\pm$0.104 &  0.213$\pm$0.051 \\
Normal               &                  0.896$\pm$0.105 &  0.567$\pm$0.257 &                0.178$\pm$0.107 &  \textbf{0.261}$\pm$0.103 \\
\midrule
\multicolumn{5}{c}{Super Mario Bros (Jump)} \\
\midrule
{} & \multicolumn{2}{c}{$\mathbb{E}[\text{playability}] \uparrow$} & \multicolumn{2}{c}{$\mathbb{E}[\text{jumps} > 0] \uparrow$} \\
\midrule
                Ours     &                  \textbf{0.990}$\pm$0.040 &  \textbf{0.995}$\pm$0.013 &                  \textbf{0.99}$\pm$0.01 &  \textbf{1.00}$\pm$0.00 \\
                Baseline &                  0.957$\pm$0.078 &  0.960$\pm$0.034 &                  0.90$\pm$0.03 &  0.75$\pm$0.08 \\
                Normal   &                  0.952$\pm$0.083 &  0.768$\pm$0.200 &                  0.90$\pm$0.02 &  0.94$\pm$0.02 \\
                \bottomrule
\bottomrule
            \end{tabular}}
        \caption{\textbf{Comparison between the discretized geometry and baselines for SMB, Zelda, and the jump submanifold}. This table shows the expected functionality (i.e. playability) and diversity of the decoded levels after running interpolations and random walks according to our proposed discretized geometry (see Sec.~\ref{sec:discretized_geometry}), as well as baselines composed of linear interpolation, Gaussian random walks and a custom center-of-mass seeking random walk. We report a mean and standard deviation after running the experiments on 10 different VAE runs for SMB, and 4 selected VAE runs for Zelda, . These results show that our proposed geometry is avoiding non-playable regions of the latent space more reliably than the baselines when performing both interpolations and random walks. However, it must be noted that this increase in reliability comes at a cost on the diversity of the decoded levels, especially when it comes to performing random walks on Zelda. This results hold even when we consider a submanifold of the original manifold, given by the levels in which Mario jumps at least once.}
        \label{table:results:comparing_geometries}
        \vspace{-0.5cm}
    \end{table}
    
    \subsection{Measuring diversity in decoded levels}

    \summaryline{Define mean similarity between levels, and then diversity as one minus similarity.} Another metric that is particularly relevant for games is the diversity present in a corpus of levels \cite{Earle_Snider_Fontaine_Nikolaidis_Togelius_2021}. We leverage the literature that discusses similarities between Categorical data \cite{Boriah_Chandola_Kumar_2008} to implement the following similarity metric: given two levels $l_1$ and $l_2$, both of size $(h,w)$, we define
    \begin{equation}
        \label{eq:similarity}
        \similarity(l_1, l_2) = \frac{1}{wh}\sum_{i,j}^{w,h} \left[l_1[i,j] = l_2[i,j]\right], 
    \end{equation}
    where $[l_1[i,j] = l_2[i,j]]$ is one when the boolean condition is satisfied, and 0 otherwise. In other words, we measure how many times the two levels $l_1$ and $l_2$ agree on average. We define the \textbf{diversity} of a corpus of levels $\mathcal{L} = \{l_m\}_{m=1}^M$ as
    \begin{equation}
        \label{eq:diversity}
        \diversity(\mathcal{L}) = 1 - \frac{2}{M(M-1)}\sum_{m < m'}\similarity(l_m, l_{m'}),
    \end{equation}
    that is, as one minus the average similarity of different pairs of levels in the corpus, counted only once. With this measure, the entire Zelda dataset used for training has a diversity of $0.23$, and the SMB dataset has a diversity of $0.17$.

    \summaryline{Results for playability on the complete grid in comparison with linear interpolation.} Table~\ref{table:results:comparing_geometries} also shows the mean diversity of the interpolations and random walks. We see that our improvements on reliablity come at a small cost on the diversity of the decoded levels. This diversity cost is stark in the case of geometric random walks for Zelda, and we hypothesize that it is because of the ``bottlenecks'' between the playable regions, as well as the fact that some playable regions may be disconnected (see Fig.~\ref{fig:jumping:our_diff} for a showcase of this behavior on similar manifolds).
    
    \subsection{A different definition of functionality}
    \label{sec:submanifolds}

    \summaryline{Idea of submanifolds, the submanifold of jumping in Mario}
    So far, we have dealt with presenting levels that are functional in the sense that they are solvable, or playable by users. We can, however, modify this definition to encompass a more restricted set. We experiment with restricting the SMB playability graph even further to the subset of levels in which Mario has to jump.
    

    This manifold is presented in Fig.~\ref{fig:jumping}, which is to be compared with Fig.~\ref{fig:model:latent_space}. Table~\ref{table:results:comparing_geometries} shows the proportion of levels in which Mario jumps in 20 interpolations and 10 random walks, just like in Sec.~\ref{sec:comparing_interpolation_random_walks}. First, we notice that the expected playability is not hindered by restricting to this submanifold. Second, since the area of levels without jump is wider than that of playabiliy (leading to a non-convex manifold), we decode to levels that include jumps more reliably than the two discussed baseline (with an estimated increase of e.g.\ 10$\%$ for interpolations).


    \section{Related work}
    
    \summaryline{All work related DGMs and videogames}
    Our work is situated among the applications of Machine Learning (ML) to Procedural Content Generation (PCG) \cite{PCGML}. More precisely, recent research has focused on applying Deep Generative Modeling to the problem of creating content for videogames. From levels in VizDoom \cite{giacomello2018doom}, graphic assets in MOBAs \cite{Karp_Swiderska-Chadaj_2021} to levels in several of the games available in the Video Game Level Corpus \cite{VGLC} for applications such as latent space exploration and evolution \cite{MarioGAN,sarkar2021generating,Earle_Snider_Fontaine_Nikolaidis_Togelius_2021}, or game blending \cite{sarkar2020conditional,sarkar2020sequential,sarkar2021dungeon,sarkar2021generating}. This research spans applications of Generative Adversarial Networks \cite{awiszus2020toadgan}, autorregresive models \cite{Summerville_Mateas_2016}, Variational Autoencoders \cite{sarkar2020conditional} and Neural Cellular Automata \cite{Earle_Snider_Fontaine_Nikolaidis_Togelius_2021}. The method we propose here differs from most of this research, given that our focus is to understand parts of the latent space that obey certain functionality criteria, and use them to define geometric algorithms for interpolation and diffusion. There exist, however, approaches in which functional content creation is addressed: \cite{torrado2019bootstrapping} bootstrap the creation of playable content; and in \cite{level_repair} authors create content and then repair it using linear programming; both approaches are complementary to ours.
    
    \summaryline{Work on UQ in VAEs and other GMs}
    We apply techniques from the Uncertainty Quantification community. Namely, the geometric quantities we discussed (e.g.\ manifolds and metrics) provide invariants to reparametrization, and thus constitute sensible tools for understanding representations \cite{hauberg:only:2018}. These geometric methods have been studied for latent variable methods like Gaussian Process Latent Variable Models (GPLVMs) \cite{Tosi:UAI:2014}, Generative Adversarial Networks \cite{Wang_Ponce_2021} and Variational Autoencoders \cite{arvanitidis:iclr:2018,arvanitidis:aistats:2019,detlefsens:proteins:2020,many:arxiv:2021}. Our contribution to this corpus consists of a different way to ensure high volume in the latent space of a Categorical VAE by calibrating a hierarchical layer in the decoding process, as well as a discretized approximation that allows for sensible random walks.
    These differential geometry tools have also been applied e.g.\  in robot control \cite{chen:metrics:2018,hadi:rss:2021,Jaquier20}. This line of research is also similar in spirit to learning functions in the latent space of e.g.\ proteins, modeled as tokens using VAEs \cite{detlefsens:proteins:2020,SchwalbeKoda:2020:chemical}.

    \section{Conclusion}

    \summaryline{We provide tools for interpolation and diffusion in the latent space of VAEs that respect functional parts, and apply them in the context of videogame levels of SMB.} In this paper we defined a geometry in the latent space of two tile-based games: Super Mario Bros and The Legend of Zelda. After learning a latent representation using a one-layer-hierarchical Variational Autoencoder (VAE), we explore the latent space for non-functional regions (e.g. unsolvable or incoherent levels) and place high metric volume there. This geometry is then approximated with a finite graph by thresholding volume, and is subsequently used for interpolations and random walks.

    \summaryline{Our proposed method reliably presents functional and diverse content, in comparison to simpler and naïve baselines.}
    When compared to simpler baselines (e.g.\ linear interpolation and Gaussian random walks), our method presents functional levels more reliably. We also tested the ability to define and explore submanifolds of the SMB playability manifold, rendering a system for computing interpolations and random walks that stay not only within playable levels, but also levels in which Mario has to jump.

    \summaryline{Future work includes scaling this up to bigger latent spaces, making this algorithms not as hyperparameter-dependent (see the scale in the diffusion)}
    There are multiple avenues for future work. First of all, many of our constructions are sensitive to the hyperparameters specified (e.g.\ the bandwith $\beta$ in Eq.~(\ref{eq:translated_sigmoid}), or the threshold cut for the metric volume in Sec.~\ref{sec:discretized_geometry}); further work is required to generalize these methods beyond the need for these hyperparameters. Moreover, the induced geometries in latent spaces have mostly been studied when the dimension of the latent space is $2$. Testing these tools and improving on them for dimensions bigger than $2$ should be addressed in future work. We also rely on the assumption that functionality is distributed smoothly in the latent space, for which we have no theoretical guarantees. Finally, learning these discrete approximations of manifolds using graphs may allow for regression methods therein \cite{borovitskiy2021matern}, opening the doors for e.g.\ optimizing certain metrics while reliably presenting functional content.
    
    \section*{Acknowledgements}
    This work was funded by a Sapere Aude: DFF-Starting Grant (9063-00046B), the Danish Ministry of Education and Science, Digital Pilot Hub and Skylab Digital. SH was supported by research grants (15334, 42062) from VILLUM FONDEN. This project has also received funding from the European Research Council (ERC) under the European Union's Horizon 2020 research and innovation programme (grant agreement Nr. 757360). This work was funded in part by the Novo Nordisk Foundation through the Center for Basic Machine Learning Research in Life Science (NNF20OC0062606).

    \bibliographystyle{IEEEtran}
    \bibliography{biblio.bib}

\end{document}